\documentclass[letterpaper]{article} % DO NOT CHANGE THIS
\usepackage{aaai2026}  % DO NOT CHANGE THIS
\usepackage{amsmath}
\usepackage{times}  % DO NOT CHANGE THIS
\usepackage{helvet}  % DO NOT CHANGE THIS
\usepackage{courier}  % DO NOT CHANGE THIS
\usepackage[hyphens]{url}  % DO NOT CHANGE THIS
\usepackage{graphicx} % DO NOT CHANGE THIS
\def\UrlFont{\rm}  % DO NOT CHANGE THIS
\usepackage{natbib}  % DO NOT CHANGE THIS AND DO NOT ADD ANY OPTIONS TO IT
\usepackage{caption} % DO NOT CHANGE THIS AND DO NOT ADD ANY OPTIONS TO IT
\usepackage{algorithm}
\usepackage{algorithmic}
\usepackage{booktabs}
\usepackage{multirow}
\usepackage{siunitx}

\usepackage[dvipsnames]{xcolor}

\usepackage{newfloat}
\usepackage{listings}
\usepackage{mathtools}
\DeclareCaptionStyle{ruled}{labelfont=normalfont,labelsep=colon,strut=off} % DO NOT CHANGE THIS
\floatstyle{ruled}
\newfloat{listing}{tb}{lst}{}
\floatname{listing}{Listing}
\title{GFocal: A Global-Focal Neural Operator for Solving PDEs on Arbitrary Geometries}
\author{
    Fangzhi Fei\textsuperscript{\rm 1},
    Jiaxin Hu\textsuperscript{\rm 1},
    Qiaofeng Li\textsuperscript{\rm 1,2,*},
    Zhenyu Liu\textsuperscript{\rm 1,3,*}
}
\affiliations{
    \textsuperscript{\rm 1}School of Mechanical Engineering, Zhejiang University, Hangzhou, China\\
    \textsuperscript{\rm 2}State Key Laboratory of Fluid Power and Mechatronic Systems, Zhejiang University, Hangzhou, China\\
    \textsuperscript{\rm 3}State Key Laboratory of Computer-Aided Design and Computer Graphics, Zhejiang University, Hangzhou, China\\
}

\begin{document}

\maketitle

\begin{abstract}
Transformer-based neural operators have emerged as promising surrogate solvers for partial differential equations, by leveraging the effectiveness of Transformers for capturing long-range dependencies and global correlations, profoundly proven in language modeling. However, existing methodologies overlook the coordinated learning of interdependencies between local physical details and global features, which are essential for tackling multiscale problems, preserving physical consistency and numerical stability in long-term rollouts, and accurately capturing transitional dynamics. In this work, we propose GFocal, a Transformer-based neural operator method that enforces simultaneous global and local feature learning and fusion.
Global correlations and local features are harnessed through Nystr\"{o}m attention-based \textbf{g}lobal blocks and slices-based
\textbf{focal} blocks to generate physics-aware tokens, subsequently modulated and integrated via convolution-based gating blocks, enabling dynamic fusion of multiscale information. GFocal achieves accurate modeling and prediction of physical features given arbitrary geometries and initial conditions. 
Experiments show that GFocal achieves state-of-the-art performance with an average 15.2\% relative gain in five out of six benchmarks and also excels in industry-scale simulations such as aerodynamics simulation of automotives and airfoils.
\end{abstract}

\section{Introduction}

Substantial significant real-world applications, such as weather forecasting \cite{fisher2009data}, structural topology optimization \cite{giannone2023aligning}, and airfoil design~\cite{chen2021mo}, require solving partial differential equations (PDEs). In practical scenarios, analytical solutions of PDEs are generally infeasible in consequence of complex geometries and non-ideal boundary conditions. Numerical approaches, such as finite element method, finite volume method, and spectral method, have been developed across diverse scenarios. However, these conventional numerical methods are computationally intensive and time-consuming, limiting their applicability in tasks where real-time evaluation or repeated functional queries are needed.%, such as online flow field reconstruction or design optimization.

%Partial Differential Equations(PDEs) serve as fundamental mathematical frameworks for modeling multiscale physical phenomena, such as weather forecasting \cite{fisher2009data}, industrial design \cite{giannone2023aligning}, and fluid dynamics \cite{gao2024generative}. While conventional numerical approaches like finite element, finite volume, and spectral methods provide solutions to these equations, they often incur high computational costs and exhibit limited adaptability to arbitrary resolutions and geometries. This computational bottleneck impedes the simulation of PDEs, particularly the prediction of large-scale transient flow.

Neural networks, leveraging rapid post-training evaluation on acceleration hardware, have emerged as a promising alternative numerical approach to solve PDEs.
Physics-informed neural networks, PINNs \cite{raissi2019physics}, formulate solving PDEs as an equivalent neural network training problem by integrating PDE residuals into the neural network loss function.
However, they suffer from computationally intensive training procedures and exhibit limited transferability to problems with similar but non-identical configurations. 
Neural operators learn functional mappings between input and output spaces, thus enabling the resolution of a class of problems rather than individual instances. 
Similar to neural networks, they retain the advantage of accelerated inference compared with conventional numerical methods, maintaining the capability to address large-scale PDE problems.

% adopt a purely data-driven modeling paradigm. These operators learn the functionals without any prior knowledge of the underlying PDE, relying solely on data-driven training, leading to faster inference times than the traditional methods. 

Prevailing categories of neural operators include DeepONet \cite{lu2021learning}, FNO \cite{li2020fourier}, and Transformer-based neural operators \cite{cao2021choose,li2022transformer,liu2022ht,wu2023solving,hao2023gnot,li2024scalable,xiao2023improved,wu2024transolver}. 
DeepONet, essentially neural fields, evaluates the output function at arbitrary domain coordinates, thus inherently suitable for learning PDE solutions on arbitrary geometries. However, this architecture is computationally less efficient compared with FNO. 
FNO can efficiently learn highly nonlinear and nonlocal features, and perform fast full-field predictions via the fast Fourier transform (FFT). 
% Besides, a uniform grid is naturally compatible with a discretization of PDEs, which is the basis of many important works. 
However, FNO is difficult to apply to non-uniformly distributed data, despite sustained efforts \cite{li2023fourier,lu2022comprehensive,tran2021factorized,liu2023nuno} to address this issue. 
Transformers and attention mechanism have revolutionized multiple domains, such as natural language processing \cite{brown2020language,devlin2019bert} and computer vision \cite{dosovitskiy2020image}, and have also been introduced into PDE solving. 
Recent works, such as ONO \cite{xiao2023improved} and Transolver \cite{wu2024transolver}, demonstrate that Transformer-based architectures can effectively solve PDEs on arbitrary geometries, offering superior performance compared to conventional numerical methods.

\begin{figure*}[h]
\centering
\includegraphics[width=0.9\textwidth,      % 宽度缩放到文本宽度的 80%
  height=0.9\textheight,    % 高度同步缩放
  keepaspectratio,          % 保持原始宽高比
  draft=false,              % 确保实际图像显示
  clip=true,                % 裁剪多余边缘
  trim=0 0 0 0              % 移除多余空白
]{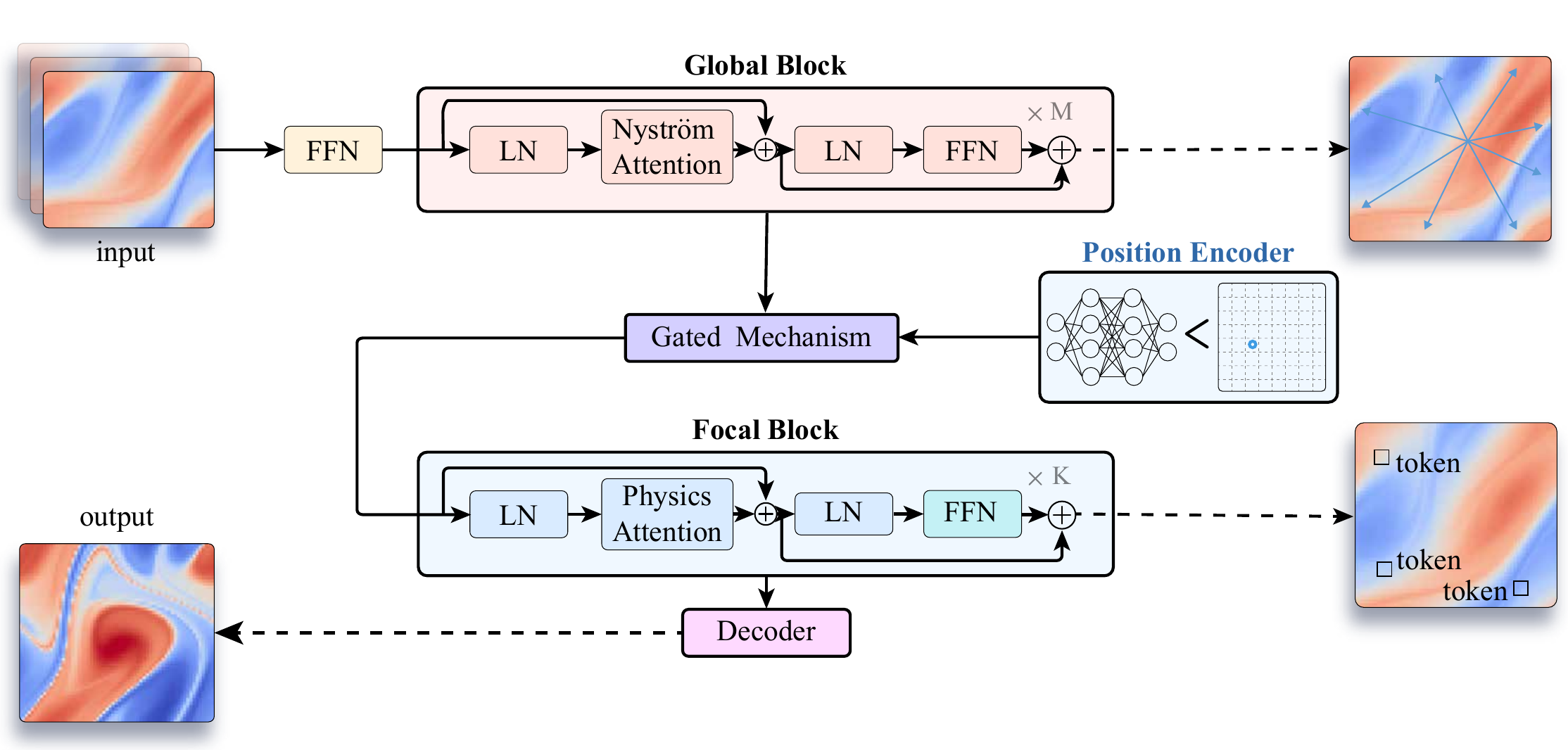} 
\caption{Overview of the GFocal architecture, consisted of 4 blocks: Global Block, Gated Mechanism, Position Encoder, and Focal Block. A feedforward network first extracts expressive features from the input data. The features are further enhanced and processed by 4 blocks and projected back into physical space by a decoder. The feature maps on the right schematically illustrate the modeling mechanisms of Global and Focal Block. The arrows indicate that the Global Block primarily focuses on modeling global information interactions; the rectangular frames indicate that the Focal Block specializes in modeling physical details by encoding different physical information into different tokens.}
\label{Fig1}
\end{figure*}

In this paper, we propose GFocal, a Global-Focal Transformer-based neural operator method for solving PDEs on arbitrary geometries. The architecture is illustrated in Figure~\ref{Fig1}. 
% We combine global and focal blocks with gated mechanism, and introduce position encoding to enhance the positional representation of features input to the focal block. 
Compared with ONO and Transolver prioritizing attention-based feature aggregation, GFocal emphasizes adaptive integration of multiscale information, critical for achieving qualitative physical fidelity in transition dynamics or enabling high-precision field reconstruction in complex systems.
GFocal learns global and local features through Nystr\"{o}m attention-based \cite{xiong2021nystromformer} global and slice-based focal blocks, adaptively modulates information flow with a convolution-based gated mechanism, and enhances the positional representation of physical features by introducing a position encoder. 
GFocal achieves consistent state-of-the-art results on five of six standard benchmarks with various geometries and two industry-scale tasks. Overall, our contributions are 
% The global block is developed to compensate for the global information in physics system. Nystr\"{o}m attention \cite{xiong2021nystromformer} is used in global block to reconstruct global feature. 
% In focal block, we use different slices to decompose the discretized domain and encapsulate inherent physical properties, while every slice focuses on learning a specific physical state by changing weights. 
% What's more, to better control the influence of features from global block, we develop gated mechanism composed of multiple convolutional layers. 

\begin{itemize}
\item We propose a modular architecture for multiscale feature extraction of PDEs by integrating Transformer-based global and focal blocks. The global block captures long-range interactions while the focal block focuses on extracting distinct physical properties within specific subregions. 
\item We introduce a Gated Mechanism enhanced by position encoding for efficient fusion of multiscale information. The gating module regulates information flow between global and focal blocks while the position encoding ensures spatial contextual awareness.
\item GFocal demonstrates superior performance, achieving an average \textbf{15.2}\% and maximum \textbf{28.3}\% relative gain in 5 out of 6 standard benchmarks.
\end{itemize}

\section{Related Work}
\noindent\textbf{Neural Operators as PDE Solvers.} To become universal PDE solvers, neural operators should generalize across diverse and unseen computational domains or geometries that are often accompanied by unstructured or non-uniform discretizations while maintaining computational efficiency. DeepONets, essentially neural fields inherently suitable for arbitrary geometries, face the challenge of slow training. FNO \cite{li2020fourier} and NUNO \cite{liu2023nuno} adopt the interpolation approach. These methods employ interpolation schemes to transform non-uniform data into uniform grids allowing for high-performance FFT processing. However, such methods introduce both interpolation and extrapolation errors, leading to degradation in computational accuracy. Others including POD-DeepONet \cite{lu2022comprehensive}, Geo-FNO \cite{li2023fourier}, GINO~\cite{li2024geometry}, and F-FNO \cite{tran2021factorized} adopt the mapping approach. They construct computational domain mappings by learning functional bijections between point clouds in arbitrary domains and uniform grids in unit domains. High performance processing is conducted in unit domains and transformed back to arbitrary domains. However, learning such mappings is difficult as handcrafted designs are needed case by case, especially for multiconnected arbitrary domains with fundamental topological difference from unit domains. GNN-based \cite{scarselli2008graph} models construct and perform computation on spatial graphs in arbitrary domains, which is essentially spatial discretization resembling finite elements or finite volumes, thus inheriting the capability of traditional physics-based solvers to handle arbitrary geometries. For example, 3D-GeoCA \cite{deng2024geometry} introduces a point cloud geometry encoder to encode the boundary of the problem domain, and conditionally guides the adaption of hidden features in the backbone model with geometry information.

\noindent\textbf{Transformer-based Neural Operators}. Transformer-based \cite{vaswani2017attention} models for PDE solving are essentially implicit discretization approaches. They directly learn the spatial and physical correlations of PDEs at discrete locations by leveraging positional encoding and attention mechanisms. Galerkin Transformer \cite{cao2021choose} as a pioneering work proposes a simplified self-attention operator without softmax normalization, achieving promising results validated both theoretically and experimentally. OFormer \cite{li2022transformer} develops a cross-attention module that enables discretization-invariant queries on output function. FactFormer \cite{li2024scalable} decomposes the input function into multiple sub-functions in 1D domain. GNOT \cite{hao2023gnot} introduces a novel heterogeneous normalized attention layer to handle multiple input functions and irregular geometries. ONO \cite{xiao2023improved} develops a novel orthogonal attention mechanism, which is inherently integrated with orthogonal regularization. These methods directly apply attention mechanisms to the original computational domain, potentially hindering the capability of the models to learn simultaneously large-scale geometric structures and fine-grained physical features.
HT-Net \cite{liu2022ht} decomposes the input-output mapping into a multi-level hierarchy and achieves feature updating through hierarchical local self-attention aggregation. While its downsampling operation enhances multiscale feature representation, the requirement for regular-shaped mesh demands error-introducing interpolation.
Transolver \cite{wu2024transolver} represents discretized computational domains through a series of learnable slices, where mesh points with physical affinity are clustered into identical slices and encoded into shared tokens. After applying attention mechanisms at the token level, these representations are reconstructed onto the original computational domain. While this approach effectively captures delicate physical details and underlying interactions, the learning of global geometry correlations remains underexplored. AMG \cite{li2024harnessing} proposes global and local sampling to construct multiscale graphs for capturing different frequencies of features, but its regional resampling step incurs substantial computational overhead.

\noindent\textbf{Attention Mechanisms}. The quadratic-complexity of the original self-attention mechanism \cite{vaswani2017attention} imposes severe scalability limits for both training and high-throughput PDE solution. More efficient attention mechanisms \cite{xiong2021nystromformer, child2019generating, cao2021choose,li2024scalable, katharopoulos2020transformers, li2022transformer, xiao2023improved, alkin2025universalphysicstransformersframework} have recently been proposed. Sparse Transformers \cite{child2019generating} employ innovative sparse factorizations of the attention matrix and reduces the computational complexity to $\mathcal{O}(n\sqrt{n})$. Nystr\"{o}mformer \cite{xiong2021nystromformer} innovatively adapts the Nystr\"{o}m method to approximate self-attention computations, achieving a reduced time complexity of $\mathcal{O}(n)$. Universal Physics Transformers \cite{alkin2025universalphysicstransformersframework} bypass iterative encoding by performing latent space temporal evolution.

\section{Method}
This section outlines the architecture of the proposed GFocal for solving PDEs on arbitrary geometries. We start with the problem setting, followed by the architecture overview. The processing modules are detailed in the following subsections.

\subsection{Problem Setup}
In this work following \cite{li2023fourier} and \cite{wu2024transolver}, we consider parametric PDEs defined on various geometric domains $D_a$, where $D_a$ is parameterized by design parameters $a \in A$. The task is to map the parameters $a$ and initial/boundary conditions $u_0$/$u_b$ to the solution $u$ through the solution operator \(\mathcal{G}^{\dagger} : (a, u_0, u_b) \mapsto u \).

\subsection{Overview of Model Architecture}
\begin{figure}[t]
\centering
\includegraphics[width=0.5\textwidth,
height=0.5\textheight,    % 高度同步缩放
  keepaspectratio,          % 保持原始宽高比
  draft=false,              % 确保实际图像显示
]{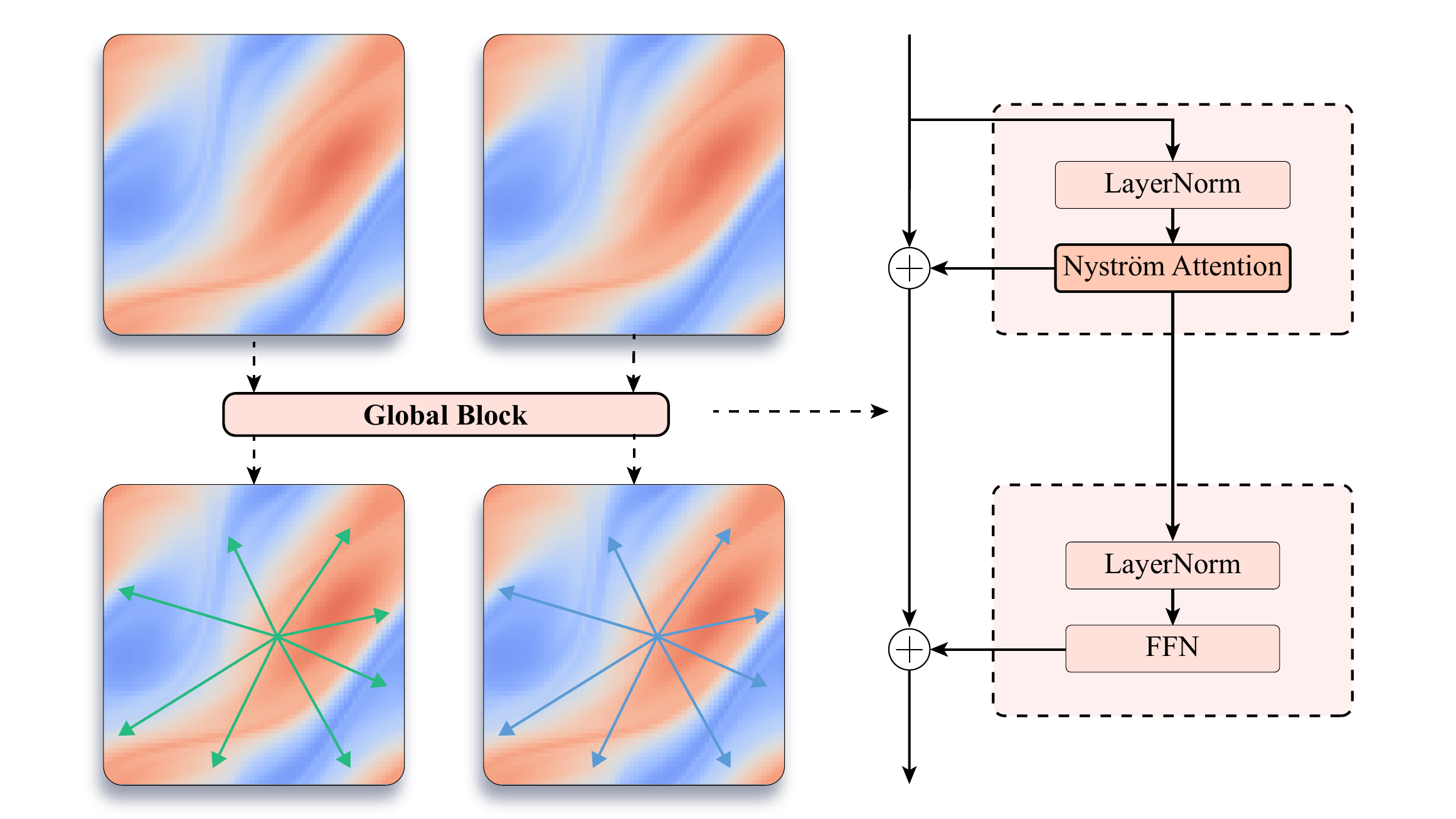}
\caption{Overview design of Global Block. We use Nystr\"{o}m attention mechanism to reconstruct global interactions.}
\label{fig2}
\end{figure}

Our model is designed to effectively capture multiscale features and correlations in PDEs on arbitrary geometries with global and focal blocks. As shown in Figure~\ref{Fig1}, our model consists of 4 components: (1) Global Block: we utilize Nystr\"{o}m attention to capture global information (shown in Figure~\ref{fig2}); (2) Position Encoder: motivated by \cite{xu2021ultrasr}, we further explore novel position encoding to enhance physical field reconstruction accuracy; (3) Gated Mechanism: this block is designed to dynamically adjust global and local information contributions; (4) Focal Block: we utilize different slices to encode different physical information into different tokens, which model local details. By applying attention mechanism to these tokens, we can effectively resolve interdependencies among distinct physical properties (shown in Figure~\ref{fig3}). See Appendix for complete proof.

\begin{figure*}[t]
\centering
\includegraphics[ width=0.65\textwidth,      % 宽度缩放到文本宽度的 80%
  height=0.65\textheight,    % 高度同步缩放
  keepaspectratio,          % 保持原始宽高比
  draft=false,              % 确保实际图像显示
  ]{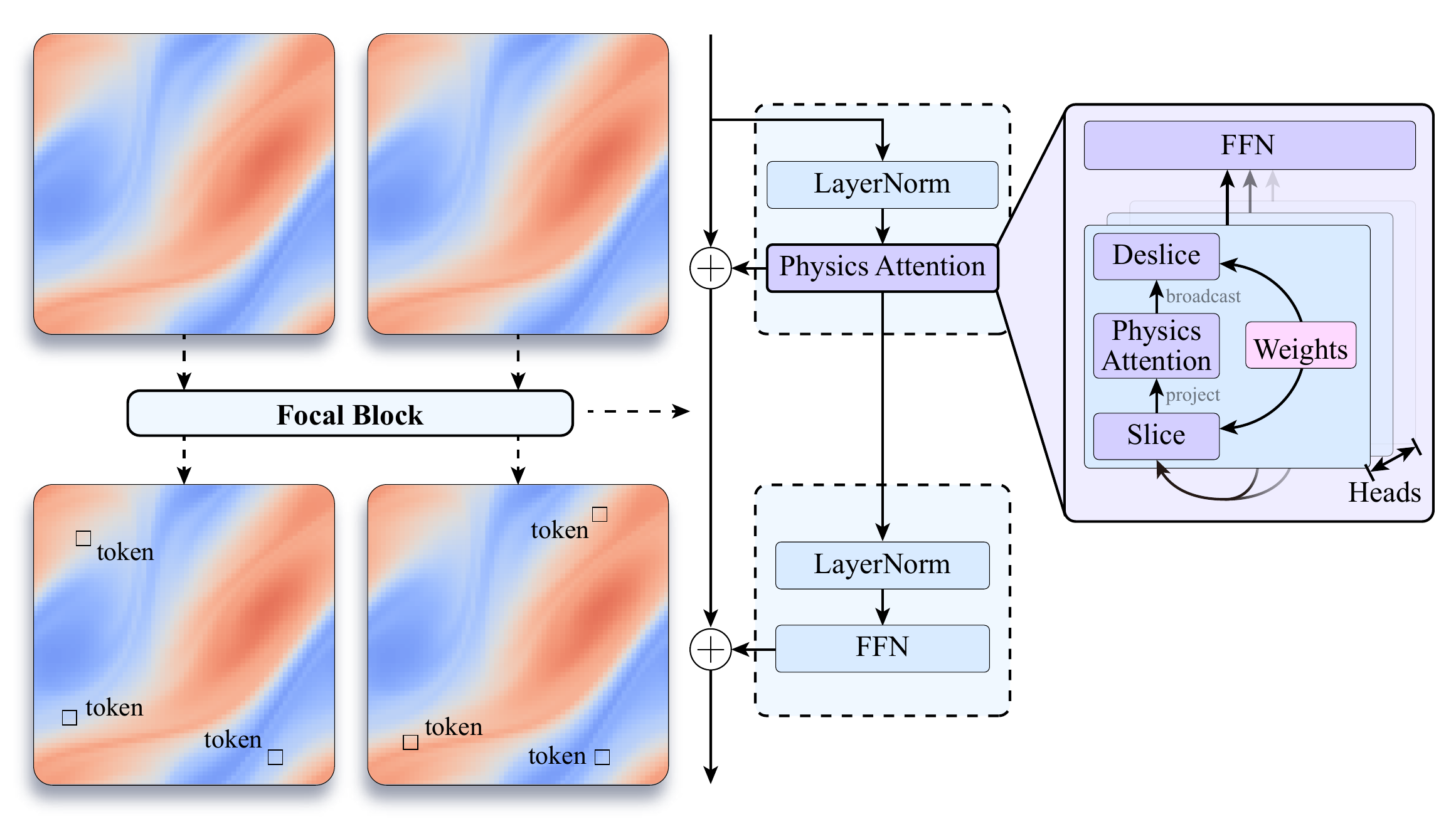} 
\caption{Overall design of Focal Block. We use a series of slices to encode information into different tokens and modeling various physical details.}
\label{fig3}
\end{figure*}

\subsection{Global Block}
Global block adopts Nystr\"{o}m attention \cite{xiong2021nystromformer} to model global information. We denote the softmax matrix used in self-attention \cite{vaswani2017attention} as $\textbf{S}$. The input information of $N$ points is encoded into feature representations {$\mathbf{F}_{\text{input}} \in \textbf{R}^{N \times C}$}, where the feature of each point contains $C$ channels. Based on these features, the softmax matrix $\textbf{S}$ can be written as $\mathbf{S} = \text{Softmax}\left( \frac{\mathbf{Q}\mathbf{K}^\top}{\sqrt{d_q}} \right) \in \textbf{R}^{N \times C}$. Nystr\"{o}m attention mechanism adopts the Nystr\"om method to approximately calculate the full softmax matrix $\textbf{S}$. The basic idea is to use landmarks $\tilde{\textbf{K}}$ and $\tilde{\textbf{Q}}$ from key $\textbf{K}$ and query $\textbf{Q}$ to derive an efficient Nystr\"om approximation without accessing the full $\textbf{QK}^\top$.  $\mathbf{S}$ is approximated as
\begin{equation}
\hat{\mathbf{S}} = 
\underbrace{\text{Softmax}\left( \frac{\mathbf{Q}\mathbf{\tilde{K}}^\top}{\sqrt{d}} \right)}_{\mathclap{N \times L}} 
\cdot
\underbrace{\mathbf{S}_{\text{a}}^{+}}_{\mathclap{L \times L}} 
\cdot
\underbrace{\text{Softmax}\left( \frac{\mathbf{\tilde{Q}}\mathbf{K}^\top}{\sqrt{d}} \right)}_{\mathclap{L \times C}}
\end{equation}
where \(\mathbf{S}_{\text{a}} = \text{Softmax} \left( \frac{\mathbf{\tilde{Q}}\mathbf{\tilde{K}}^{\top}}{\sqrt{d}} \right) \). Suppose that there are $M$ layers in the global block, the $m$-th layer can be formalized as
 \begin{equation}
	\begin{split}
\hat{\mathbf{F}}^{m} &= \hat{\mathbf{S}}\left(\operatorname{LayerNorm}\left({\mathbf{F}}^{m-1}\right)\right) + {\mathbf{F}}^{m-1}, \\
{\mathbf{F}}^{m} &= \operatorname{FeedForward}\left(\operatorname{LayerNorm}\left(\hat{\mathbf{F}}^{m}\right)\right) + \hat{\mathbf{F}}^{m}
	\end{split}
\end{equation}
where $m \in \{1, \ldots, M\}$. $\mathbf{F}^m \in \mathbf{R}^{N \times C}$ is the output of the $m$-th layer. $\mathbf{F}^0=\mathbf{F}_{\text{input}} \in \mathbf{R}^{N \times C}$ represents the input deep feature, which is encoded from input observation. We obtain the final output as $\mathbf{F}_{\text{global}}$.

\subsection{Position Encoder}
Inspired by the recent advances in image super-resolution \cite{dosovitskiy2020image}, we propose a structured approach to effectively capture spatial correlations and enhance the model performance. We define a reference grid within the domain \([0,1] \times [0,1]\) as \(\mathbf{G}_{\text{ref}} \in R^{B \times R \times R \times D}\), where $B$ is the batch size, $R$ is the resolution, $D$ is the coordinate dimension. Taking $D=2$ as an example, we uniformly sample \(R\) points in the interval \([0, 1]\)
\begin{equation}
g_x = g_y = \left\{ \frac{k}{R-1} \mid k = 0, 1, \ldots, R-1 \right\}
\end{equation}
Each point \(\mathbf{G}_{i,j}\) in the reference domain is given by
\begin{equation}
\mathbf{G}_{i,j} = \begin{pmatrix}
g_x[i] \\
g_y[j]
\end{pmatrix}, \enspace \text{where} \; i, j = 0, 1, \ldots, R - 1
\end{equation}
Given the coordinates $\mathbf{p}$ of an input point, its Euclidean distance with respect to all reference grid points $\mathbf{G}_{i,j}$ are calculated
\begin{flalign}
d(\mathbf{p}, \mathbf{G}_{i,j}) = \|\mathbf{p} - \mathbf{G}_{i,j}\|_2
\end{flalign}
The above coordinate transformation pipeline converts each input point $\mathbf{p}$ into an $R \times R$ relative coordinate representation $d(\mathbf{p}, \mathbf{G}_{i,j})$, which is fed into a vanilla MLP.

\subsection{Gated Mechanism}
A gated mechanism is designed to modulate the global and local information flow. The output of the global block  $\mathbf{F}_{\text{global}}$ is passed through convolutional layers followed by sigmoid activations to produce gating maps \(\mathbf{G} \in \mathbf{R}^{B \times N \times C}\):
\begin{flalign}
\mathbf{G} = \sigma \left( \text{Conv}\left( \mathbf{F}_{\text{global}} \right) \right)
\end{flalign}
where Conv is a convolution layer that reduces the features to a single-channel gating map, $\sigma(\cdot)$ is the element-wise sigmoid activation function. The gated mechanism block comprises 3 convolutional layers. The $g$-th layer of gating block can be formalized as
\begin{flalign}
\mathbf{F}^g = \mathbf{G} \odot \mathbf{F}^{g-1} + \textrm{MLP}(d(\mathbf{p}, \mathbf{G}_{i,j}))
\end{flalign}
where $g \in \{1, 2, 3\}$, $\odot$ denotes element-wise multiplication. The gating map $\mathbf{G}$ is broadcast across the channel dimension to match the dimensions of the feature maps. We obtain the output of the gating block as $\mathbf{F}_{\text{gated}}$, which is then passed into the focal block.

\subsection{Focal Block}
While global block excels at modeling global dependencies, it does not fully exploit local interactions critical for detailed physical field reconstruction. In \cite{wu2024transolver}, researchers found that discrete points in the solution domain essentially constitute finite samples of the underlying continuous physical space, and learning the intrinsic nature of physical states has emerged as an effective methodology for reconstructing physical fields. To characterize the physical properties of the field, we assign each point $i$ to $L$ potential slices based on its learned feature $\mathbf{f}_{i}$. The classification process for each point in the computational domain is computed as 
\begin{equation}
\{\mathbf{w}_{i}\}_{i=1}^{N} = \left\{\operatorname{Softmax}\big(\operatorname{Map}\left(\mathbf{f}_{i}\right)\big)\right\}_{i=1}^{N} 
\end{equation}
where $\operatorname{Map}()$ projects the $C$-dimensional input features into $L$ dimensions. As for structured meshes or uniform grid, $\operatorname{Map}()$ is a local convolution layer; and for unstructured meshes or non-uniform grid, a linear layer. $\mathbf{w}_{i,j}$ represents the contribution of the ${i}$-th point to the $j$-th slices with $\sum_{j=1}^{L} \mathbf{w}_{i,j}=1$. Points with similar weight vectors will be assigned to the same class.
$\mathbf{s}_{j}\in \mathbf{R}^{N\times C}$ represents the $j$-th slice feature, which is a {weighted combination of $N$ point features $\mathbf{f}_i$ and is computed as follows.
\begin{equation}
\mathbf{s}_{j} = \left\{\mathbf{w}_{i,j}\mathbf{f}_{i}\right\}_{i=1}^{N}
\end{equation}
We encode features into different tokens by spatially weighted aggregation as
\begin{equation}
	\begin{split}
\mathbf{t}_{j} = \frac{\sum_{i=1}^{N} \mathbf{s}_{j,i}}{\sum_{i=1}^{N}\mathbf{w}_{i,j}} = \frac{\sum_{i=1}^{N} \mathbf{w}_{i,j}\mathbf{f}_{i}}{\sum_{i=1}^{N} \mathbf{w}_{i,j}}
	\end{split}
\end{equation}
where physically consistent slices $\mathbf{s}=\{\mathbf{s}_{j}\}_{j=1}^{L}\in \mathbf{R}^{L\times (N\times C)}$, slice weights $\mathbf{w}\in \mathbf{R}^{N\times L}$, and physics-aware tokens $\mathbf{t}=\{\mathbf{t}_{j}\}_{j=1}^{L}\in \mathbf{R}^{L\times C}$. Then the Physics Attention Mechanism , as shown in Figure~\ref{fig3}, is applied to capture intrinsic interconnections between different physical states, which is different from geometrical correlation, and is formulated as
\begin{equation}
\begin{split}
\mathbf{q}, \mathbf{k}, \mathbf{v} = \operatorname{Linear}(\mathbf{t}), \ \ \mathbf{t}^\prime = \operatorname{Softmax}\left(\frac{\mathbf{q}\mathbf{k}^{\sf T}}{\sqrt{C}}\right)\mathbf{v}
	\end{split}
\end{equation}
where $\mathbf{q}, \mathbf{k}, \mathbf{v}, \mathbf{t}^\prime\in \mathbf{R}^{L\times C}$.
Transited physical tokens $\mathbf{t}^\prime=\{\mathbf{t}_{j}^\prime\}_{j=1}^{L}$ are then transformed back to corresponding points by deslicing
\begin{equation}
	\begin{split}
\mathbf{f}_{i}^\prime & = \sum_{j=1}^{L} \mathbf{w}_{i,j}\mathbf{t}_{j}^\prime
	\end{split}
\end{equation}
where $i=1,\dots,N$ and each token $\mathbf{t}_{j}^\prime$ is broadcast to all points. The above process is denoted for brevity as $\mathbf{f}^\prime = \operatorname{Physics-Attn}(\mathbf{f}) = \operatorname{Physics-Attn}(\left\{\mathbf{f}_i\right\}_{i=1}^{N})$. The focal block consists of $K$ Physics Attention layers, as shown in Figure~\ref{fig3}. The $k$-th layer is formulated as
\begin{equation}
	\begin{split}
\hat{\mathbf{F}}^{k} &= \operatorname{Physics-Attn}\left(\operatorname{LayerNorm}\left({\mathbf{F}}^{k-1}\right)\right) + {\mathbf{F}}^{k-1}, \\
{\mathbf{F}}^{k} &= \operatorname{FeedForward}\left(\operatorname{LayerNorm}\left(\hat{\mathbf{F}}^{k}\right)\right) + \hat{\mathbf{F}}^{k}
	\end{split}
\end{equation}
We obtain the final output as $\mathbf{F}_{\text{focal}}$. Then we feed it into decoder and get the final prediction.

\section{Experiment}
We conduct extensive experiments on diverse and challenging benchmarks across various domains to showcase the effectiveness of the proposed method.

\subsection{Benchmarks}
As shown in Table~\ref{tab1}, we conduct experiments on various PDE solution tasks, including Elasticity, Plasticity, Airfoil, Pipe, Navier-Stokes, and Darcy, spanning across various spatial data formats in both 2D and 3D, including point clouds, structured meshes, regular grids, and unstructured meshes. 
We also include industry-scale tasks AirfRANS and Shape-Net Car. AirfRANS \cite{bonnet2022airfrans} contains high-fidelity simulation data for Reynolds-Averaged Navier–Stokes equations on airfoils from the National Advisory Committee for Aeronautics. The task is to estimate the surface pressure and surrounding air velocity given airfoil shapes in 2D. The Shape-Net Car task \cite{umetani2018learning} is to estimate the surface pressure and surrounding air velocity given automotive shapes in 3D. See Appendix for comprehensive descriptions of implementations.

\begin{table}[h]
    \caption{Summary of experiment benchmarks.}
    \label{tab1}
    \centering
    \setlength{\tabcolsep}{1pt}
    \begin{tabular}{c | c | c | c}
        \toprule
        Geometry & Benchmarks & Input & Output  \\
        \midrule
        Point Cloud & Elasticity  & Structure & Stress \\
        \midrule
        Structured & Plasticity & External Force & Displacement  \\
        Mesh & Airfoil  & Structure & Mach Number \\
         & Pipe & Structure & Velocity  \\
        \midrule
        Regular  & Navier-Stokes & Velocity  &  Velocity  \\     
        Grid & Darcy  &  Porous Medium  & Pressure \\
        \midrule
        Unstructured & Shape-Net Car & Structure & Velocity \&  \\
        Mesh & AirfRANS &  Structure & Pressure   \\
        \bottomrule
    \end{tabular}
\end{table}

\subsection{Baselines}
We comprehensively compare GFocal with more than 14 baselines, including Geo-FNO \cite{li2023fourier}, U-NO \cite{rahman2022u}, LSM \cite{wu2023solving}, GNOT \cite{hao2023gnot}, FactFormer \cite{li2024scalable}, and Transolver \cite{wu2024transolver}, etc. 

\begin{table*}[h]
	\caption{Performance comparison on standard benchmarks. Relative L2 Loss is recorded. A smaller value indicates better performance. For clarity, the best result is in bold and the second best is underlined. Promotion refers to the relative error reduction w.r.t. the second best model ($1-\frac{\text{GFocal error}}{\text{The second best error}}$) on each benchmark. ``*'' indicates that the results are reproduced by ourselves. ``/'' means that the baseline cannot apply to this benchmark. ``-'' means that GFocal does not achieve the best performance in this benchmark.}
	\label{tab2}
	\centering
		\begin{sc}
			\renewcommand{\multirowsetup}{\centering}
			\begin{tabular}{l|cccccc}
				\toprule
                    \multirow{3}{*}{Model} & \multicolumn{1}{c}{Point Cloud} & \multicolumn{3}{c}{Structured Mesh} & \multicolumn{2}{c}{Regular Grid} \\
                    \cmidrule(lr){2-2}\cmidrule(lr){3-5}\cmidrule(lr){6-7}
				& Elasticity & Plasticity & Airfoil & Pipe & Navier–Stokes & Darcy \\
				\midrule
                    WMT& 0.0359 & 0.0076 & 0.0075 & 0.0077 & {0.1541} & 0.0082 \\
                    U-FNO&0.0239 & 0.0039 & 0.0269 & {0.0056} & 0.2231 & 0.0183 \\
                    geo-FNO& {0.0229} & 0.0074 & 0.0138 & 0.0067 & 0.1556 & 0.0108 \\
                    U-NO& 0.0258 & {0.0034} & 0.0078 & 0.0100 & 0.1713 & 0.0113 \\
                    F-FNO&0.0263 & 0.0047 & 0.0078 & 0.0070 & 0.2322 & {0.0077} \\
                    LSM& 0.0218 & 0.0025 & 0.0059 & 0.0050 & 0.1535 & 0.0065 \\
                    Galerkin& 0.0240 & 0.0120 & 0.0118 & 0.0098 & 0.1401 & 0.0084 \\
                    \midrule
                    HT-Net& / & 0.0333 & 0.0065 & 0.0059 & 0.1847 & 0.0079 \\
                    OFormer
                    & 0.0183 & 0.0017 & 0.0183 & 0.0168 & 0.1705 & 0.0124 \\
                    GNOT& 0.0086 & 0.0336 & 0.0076 & 0.0047 & 0.1380 & 0.0105 \\
                    FactFormer& / & 0.0312 & 0.0071 & 0.0060 & 0.1214 & 0.0109 \\
                    ONO & 0.0118 & 0.0048 & 0.0061 & 0.0052 & 0.1195 & 0.0076 \\
                    Transolver*&  \underline{0.0060} & \underline{0.0012} & \underline{0.0055} & \underline{0.0046} & \underline{0.0882} & \textbf{0.0052}\\
                    \midrule
                    \textbf{GFocal (Ours)} & \textbf{0.0043} & \textbf{0.0012} & \textbf{0.0048} & \textbf{0.0035} & \textbf{0.0784} & \underline{0.0057} \\
                    Relative Promotion & 28.3\% & 0\% & 12.7\% & 23.9\% & 11.1\% & - \\
				\bottomrule
			\end{tabular}
		\end{sc}
\end{table*}

\subsection{Implementation}
To ensure fair performance comparison, all experiments were conducted on a single Nvidia RTX A6000 Ada GPU and default hyperparameters were employed for the baseline models. See Appendix for comprehensive descriptions of implementations.

\subsection{Main Results}
\noindent\textbf{Standard Benchmarks}. As presented in Table~\ref{tab2}, the proposed GFocal achieves the lowest errors in 5 out of 6 widely-used benchmarks, covering solid and fluid physics in various input data formats, demonstrating its effectiveness in handling complex geometries. As shown in Figure~\ref{fig4}, GFocal gains significant promotion in point cloud and structured mesh tasks (28.3\% in Elasticity, 23.9\% in Pipe, and 12.7\% in Airfoil).
Previous advanced Transformer-based models, such as Transolver, underemphasize the coordinated learning of local physical details and global correlations, performing suboptimal in the Elasticity task.
In the case of NS2d and Plasticity, which involve temporal predictions, our model surpasses all baselines, highlighting its temporal generalization capability.
GFocal achieves the second-lowest prediction error on the Darcy benchmark, with a slight margin compared to the best model. 

\begin{figure*}[h]
\centering
\includegraphics[
width=\textwidth,      % 宽度缩放到文本宽度的 80%
%  height=0.8\textheight,    % 高度同步缩放
 % keepaspectratio,          % 保持原始宽高比
  %draft=false,              % 确保实际图像显示
]{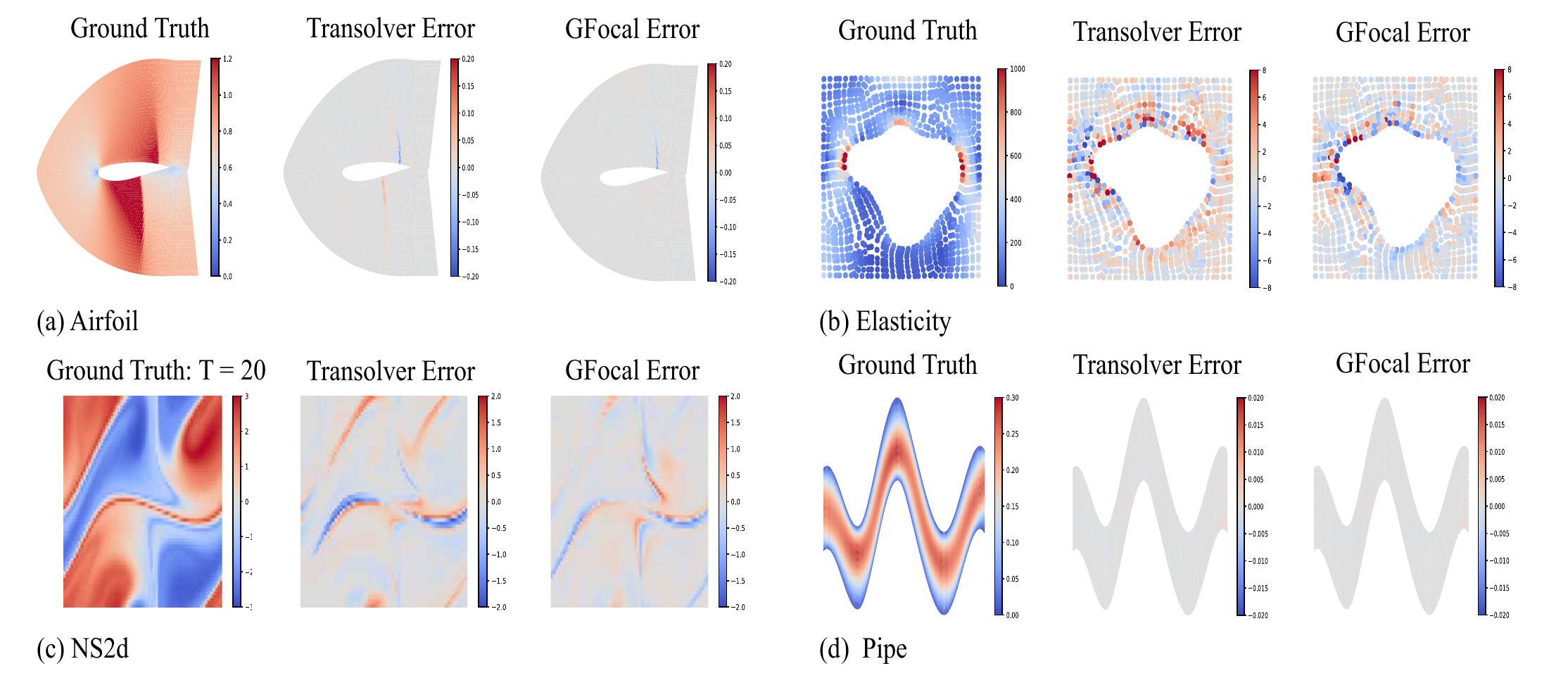} 
\caption{Error map comparison between the state-of-the-art model Transolver and the proposed GFocal.}
\label{fig4}
\end{figure*}

\begin{table}[h]
\caption{Performance on AirfRANS dataset. The relative L2 of the surrounding (Volume) and surface (Surf) physics fields as well as the relative L2 of lift coefficient ($C_{L}$) are recorded. The Spearman's rank correlation coefficients $\rho_{L}$ approaching 1 indicates better model performance.}
\label{tab3}
\centering
\begin{tabular}{l|cccc}
\toprule
Model & Volume $\downarrow$ & Surf $\downarrow$ & $C_{L}$ $\downarrow$ & $\rho_{L}$ $\uparrow$ \\
\midrule
Simple MLP & 0.0081 & 0.0200 & 0.2108 & 0.9932 \\
MeshGraphNet & 0.0214 & 0.0387 & 0.2252 & 0.9945 \\
GNO& 0.0269 & 0.0405 & 0.2016 & 0.9938 \\
Galerkin & 0.0074 & 0.0159 & 0.2336 & 0.9951 \\
\midrule
Geo-FNO& 0.0361 & 0.0301 & 0.6161 & 0.9257 \\
GNOT& 0.0049 & 0.0152 & 0.1992 & 0.9942 \\
GINO& 0.0297 & 0.0482 & 0.1821 & 0.9958 \\
3D-GeoCA & / & / & / & / \\
Transolver & \textbf{0.0037} & \underline{0.0142} & \underline{0.1030} & \underline{0.9978} \\
\midrule
\textbf{GFocal (Ours)} & \underline{0.0041} & \textbf{0.0061} & \textbf{0.0894} & \textbf{0.9980} \\
\bottomrule
\end{tabular}
\end{table}

\noindent\textbf{Large-scale tasks}. To validate the effectiveness of GFocal in complex industry-scale applications, we also explore two CFD datasets: AirfRANS \cite{bonnet2022airfrans} and Shape-Net Car \cite{umetani2018learning} with unstructured mesh.

\begin{table}[h]
\centering
\caption{Performance on Shape-Net Car dataset. The relative L2 of the surrounding (Volume) and surface (Surf) physics fields as well as drag coefficient ($C_{D}$) are recorded. The Spearman's rank correlation coefficients $\rho_{D}$ approaching 1 indicates better model performance.}
\label{tab:shapenet}
\begin{tabular}{l|cccc}
\toprule
Model & Volume $\downarrow$ & Surf $\downarrow$ & $C_{D}$ $\downarrow$ & $\rho_{D}$ $\uparrow$ \\
\midrule
Simple MLP & 0.0512 & 0.1304 & 0.0307 & 0.9496 \\
MeshGraphNet & 0.0354 & 0.0781 & 0.0168 & 0.9840 \\
GNO& 0.0383 & 0.0815 & 0.0172 & 0.9834 \\
Galerkin& 0.0339 & 0.0878 & 0.0179 & 0.9764 \\
\midrule
Geo-FNO & 0.1670 & 0.2378 & 0.0664 & 0.8280 \\
GNOT & 0.0329 & 0.0798 & 0.0178 & 0.9833 \\
GINO & 0.0386 & 0.0810 & 0.0184 & 0.9826 \\
3D-GeoCA& 0.0319 & \underline{0.0779} & 0.0159 & 0.9842 \\
Transolver& \textbf{0.0207} & \textbf{0.0745} & \underline{0.0103} & \underline{0.9935} \\
\midrule
\textbf{GFocal (Ours)} & \underline{0.0232} & 0.0792 & \textbf{0.0096} & \textbf{0.9945} \\
\bottomrule
\end{tabular}
\end{table}

As shown in Table~\ref{tab1}, the tasks are to estimate the surface and surrounding physical fields given complex geometries. As shown in Table~\ref{tab3} and~\ref{tab:shapenet}, the results demonstrate that GFocal also achieves outstanding performance in these two complex tasks compared to other models. Most notably, GFocal performs best in design-oriented metrics, including drag and lift coefficients, as well as Spearman’s rank correlation. Spearman's rank correlation quantifies the capability of the model to rank diverse shape designs correctly despite the existence of fitting errors. A model with a high Spearman's rank correlation accurately quantifies the relative superiority of design alternatives, thereby enabling effective design optimization through reliable performance differentiation.
GFocal also demonstrates superior accuracy in predicting physical fields on airfoil surfaces, although its performance in capturing both the surrounding flow fields of airfoils and the surface/surrounding physical fields of car models exhibits marginally reduced performance compared with Transolver.
\begin{figure}[h]
\centering
\includegraphics[
width=0.35\textwidth,      % 宽度缩放到文本宽度的 80%
  height=0.35\textheight,    % 高度同步缩放
  keepaspectratio,          % 保持原始宽高比
  draft=false,              % 确保实际图像显示
]{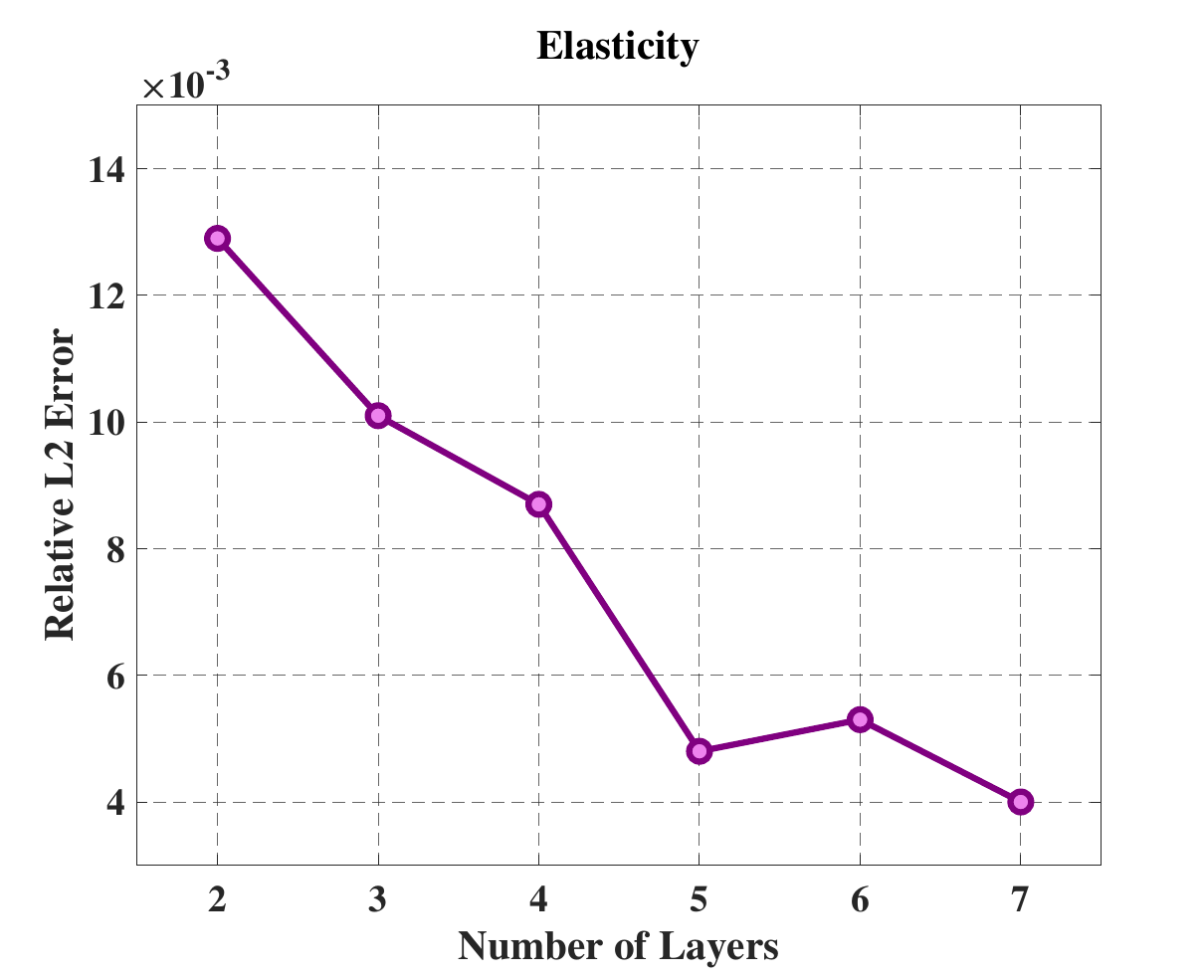}
\caption{Scaling experiment results for different numbers of layers. We keep the same number of layers for global and focal blocks.}
\label{fig5}
\end{figure}
\subsection{Model Analysis}
\noindent\textbf{Ablation Study.} We conducted ablation studies to determine the contribution of different components to the performance of GFocal. The results are summarized in Table~\ref{tab4}. As expected GFocal outperforms all ablated variants. ``w/o Focal'' confirms the necessity of learning local physical states; ``w/o Gate'' confirms the necessity of effectively fusing global and local information; ``w/o PE'' shows that positional encoders capture spatial correlations in PDEs and enhance the model performance. 

\begin{table}[h]
\centering
\caption{Ablation study on GFocal. ``w/o'' indicates the removal of this block from the original architecture.}
\label{tab4}
 \setlength{\tabcolsep}{2.5pt}
\begin{tabular}{l | cccc}
\toprule
\textbf{Case} & \textbf{w/o Focal} & \textbf{w/o Gate} & \textbf{w/o PE} & \textbf{GFocal} \\ 
\midrule
ELASTICITY      & 0.0052          & 0.0045          & 0.0043          & 0.0043    \\
AIRFOIL   & 0.0052          & 0.0050          & 0.0052          & 0.0048       \\
\bottomrule
\end{tabular}
\end{table}

\noindent\textbf{Scaling Experiments.} We conduct experiments to validate the scaling capability of GFocal. The results in Figure~\ref{fig5} demonstrate a positive correlation between error reduction and layer depth. Thus, we adopt a 5-layer architecture that delivers optimal performance with manageable computational cost. Additional results on scaling experiments and the analysis of model efficiency are provided in Appendix.

\section{Conclusions}
We propose GFocal, a Transformer-based neural operator method, to solve PDEs on arbitrary geometries. 
GFocal integrates Nystr\"{o}m attention-based global blocks and Physics attention-based focal blocks through convolution-based gated mechanism, assisted by positional encoding. 
GFocal outperforms existing state-of-the-art methods on the majority of metrics in five out of six widely-adopted benchmarks, with an average 15.2\% and maximum 28.3\% relative gain, and two industry-scale engineering tasks. 
The results demonstrate that the multi-level information fusion architecture of GFocal enables effective geometric feature extraction and superior modeling of complex fluid-structure interactions.
Future directions include extending the model to broader industrial applications and investigating its feasibility as a pre-trained foundation model for multi-physics PDE systems.

\section{Acknowledgement}
We would like to thank Yilei Qin and Zituo Chen for constructive advice. The paper is supported by the Major Research Plan of the National Natural Science Foundation of China (Grant No. 92470109) and Zhejiang Provincial Team of Leading Talents in Innovation and Entrepreneurship (Grant No. 2024R01002).

\bibliography{aaai2026}
\clearpage % 确保内容刷新
\onecolumn % 关键命令：切换单栏模式
\appendix % 开始附录编号
\makeatletter
\@twocolumnfalse
\makeatother
\urlstyle{rm} % DO NOT CHANGE THIS
\def\UrlFont{\rm}  % DO NOT CHANGE THIS
\frenchspacing  % DO NOT CHANGE THIS
\setlength{\pdfpagewidth}{8.5in}  % DO NOT CHANGE THIS
\setlength{\pdfpageheight}{11in}  % DO NOT CHANGE THIS
%
% These are recommended to typeset algorithms but not required. See the subsubsection on algorithms. Remove them if you don't have algorithms in your paper.
%\usepackage{algorithm}
%\usepackage{algorithmic}
%\usepackage{booktabs}
%\usepackage{multirow}
%\usepackage{siunitx}

%%%% adding commands
%\usepackage[usernames,dvipsnames]{xcolor}
% \newcommand{\planning}[1]{\noindent   {\colorbox{Mahogany}{\color{White}
%  PLANNING:} \color{Bittersweet} #1 \normalcolor}}
% \newcommand{\continue}[1]{\noindent  {\colorbox{RoyalBlue} {\color{White}
% CONTINUE HERE:} \color{RoyalBlue} #1 \normalcolor}}
% \newcommand{\revise}[1]{\noindent  {\colorbox{Fuchsia}{\color{White} 
% REVISE:} \color{Fuchsia}  #1 \normalcolor}}
% \newcommand{\flushout}[1]{\noindent {\colorbox{OliveGreen}{\color{White} 
% FLUSH-OUT:} \color{OliveGreen} #1 \normalcolor}}
% \newcommand{\fix}[1]{\noindent  {\colorbox{BrickRed}{\color{White} 
% FIX:}  \color{BrickRed} #1 \normalcolor}}
% \newcommand{\alert}[1]{\noindent  {\colorbox{BrickRed}{\color{White} 
% ALERT!:}  \color{BrickRed} #1 \normalcolor}}
% \newcommand{\note}[1]{\noindent  {\colorbox{Goldenrod}{\color{White} 
% NOTE:}  \color{Goldenrod} #1 \normalcolor}}
% \newcommand{\addref}[1]{\noindent  {\colorbox{blue}{\color{White} 
% ADDREF:}  \color{blue} #1 \normalcolor}}
\newcommand{\fangzhi}[1]{\noindent  {\colorbox{Periwinkle}{\color{White} 
Fangzhi plz write here:}  \color{Periwinkle} #1\normalcolor}}
\newcommand{\fangzhifix}[1]{\noindent  {\colorbox{BrickRed}{\color{White} 
Fangzhi, plz fix:}  \color{BrickRed} #1 \normalcolor}}

%
% These are are recommended to typeset listings but not required. See the subsubsection on listing. Remove this block if you don't have listings in your paper.
%\usepackage{newfloat}
%\usepackage{listings}
%\DeclareCaptionStyle{ruled}{labelfont=normalfont,labelsep=colon,strut=off} % DO NOT CHANGE THIS
\lstset{%
	basicstyle={\footnotesize\ttfamily},% footnotesize acceptable for monospace
	numbers=left,numberstyle=\footnotesize,xleftmargin=2em,% show line numbers, remove this entire line if you don't want the numbers.
	aboveskip=0pt,belowskip=0pt,%
	showstringspaces=false,tabsize=2,breaklines=true}
\floatstyle{ruled}
\newfloat{listing}{tb}{lst}{}
\floatname{listing}{Listing}
%
% Keep the \pdfinfo as shown here. There's no need
% for you to add the /Title and /Author tags.
\pdfinfo{
/TemplateVersion (2026.1)
}

% \begin{document}
\section{Proof of Theorem}\label{appendix:proof}
This proof is to demonstrate that the attention mechanism employed in the GFocal can be formalized as a Monte-Carlo approximation of an integral operator.
Given input function $\boldsymbol{a}:\Omega\to \mathbf{R}^{C}$, the integral operation $\mathcal{G}$ is
\begin{equation}
	\begin{split}\label{equ:integral_form}
 \mathcal{G}(\boldsymbol{a})(\mathbf{x})=\int_{\Omega}\kappa(\mathbf{x}, \boldsymbol{\xi})\boldsymbol{a}(\boldsymbol{\xi})\mathrm{d}\boldsymbol{\xi}
    \end{split}
\end{equation}
where $\mathbf{x}\in\Omega\subset \mathbf{R}^{C}$ and $\kappa(\cdot,\cdot)$ denotes the kernel function defined on $\Omega$. According to the formalization of attention mechanism, we propose to define the kernel function as
\begin{equation}
	\begin{split}\label{equ:kernel}
\kappa(\mathbf{x}, \boldsymbol{\xi})= \left(\int_{\Omega}\exp \left(\left(\mathbf{W}_{\mathbf{q}}\boldsymbol{a}(\boldsymbol{\xi}^\prime)\right) \left(\mathbf{W}_{\mathbf{k}}\boldsymbol{a}(\boldsymbol{\xi})\right)^{\sf T}\right)\mathrm{d}\boldsymbol{\xi}^\prime\right)^{-1}\exp \left(\left(\mathbf{W}_{\mathbf{q}}\boldsymbol{a}(\mathbf{x})\right) \left(\mathbf{W}_{\mathbf{k}}\boldsymbol{a}(\boldsymbol{\xi})\right)^{\sf T}\right)\mathbf{W}_{\mathbf{v}}
    \end{split}
\end{equation}
where $\mathbf{W}_{\mathbf{q}}, \mathbf{W}_{\mathbf{k}}, \mathbf{W}_{\mathbf{v}}\in \mathbf{R}^{C\times C}$. Softmax normalizing function is used across domain $\Omega$, and we assume that $\Omega$ carries a uniform measure for simplicity in the Monte-Carlo approximation. Supposing there are $N$ discretized points $\{\mathbf{x}_{1},\cdots,\mathbf{x}_{N}\}$, where $\mathbf{x}_{i}\in\Omega\subset \mathbf{R}^{C}$, Equation~\eqref{equ:kernel} can be approximated as
\begin{equation}
\label{equ:approximation}
	\begin{split}
\int_{\Omega}\exp \left(\left(\mathbf{W}_{\mathbf{q}}\boldsymbol{a}(\boldsymbol{\xi}^\prime)\right) \left(\mathbf{W}_{\mathbf{k}}\boldsymbol{a}(\boldsymbol{\xi})\right)^{\sf T}\right)\mathrm{d}\boldsymbol{\xi}^\prime
 \approx \frac{|\Omega|}{N} \sum_{i=1}^{N}\exp \left(\left(\mathbf{W}_{\mathbf{q}}\boldsymbol{a}(\mathbf{x}_{i})\right) \left(\mathbf{W}_{\mathbf{k}}\boldsymbol{a}(\boldsymbol{\xi})\right)^{\sf T}\right)
    \end{split}
\end{equation}
Applying Eq.~\eqref{equ:approximation} to Eq.~\eqref{equ:integral_form}, $ \mathcal{G}(\boldsymbol{a})(\mathbf{x})$ can be described as a weighted sum
\begin{equation}
	\begin{split}\label{equ:final}
 \mathcal{G}(\boldsymbol{a})(\mathbf{x})\approx\sum_{i=1}^{N}\frac{\exp \left(\left(\mathbf{W}_{\mathbf{q}}\boldsymbol{a}(x)\right) \left(\mathbf{W}_{\mathbf{k}}\boldsymbol{a}(\mathbf{x}_{i})\right)^{\sf T}\right)\mathbf{W}_{\mathbf{v}}\boldsymbol{a}(\mathbf{x}_{i})}{ \sum_{j=1}^{N}\exp \left(\left(\mathbf{W}_{\mathbf{q}}\boldsymbol{a}(\mathbf{x}_{j})\right) \left(\mathbf{W}_{\mathbf{k}}\boldsymbol{a}(\mathbf{x}_{i})\right)^{\sf T}\right)}
    \end{split}
\end{equation}
which demonstrates the equivalence of the attention mechanism to a learnable integral operator in the solution domain.

\section{Details For Benchmarks}\label{appendix:benckmark detail}
% We evalute GFocal across eight benchmarks, which involve three types of PDEs: Solid material, Navier-Stokes equations for fluid and Darcy's law. Here are the details of each benchmarks.
\subsection{Elasticity}
The governing equation of a solid body is given by
\begin{equation}
\rho^s \frac{\partial^2 u}{\partial t^2} + \nabla \cdot \sigma = 0
\end{equation}
where \(\rho^s\) is the density, \(u\) is the displacement vector, \(\sigma\) is the stress tensor. We study a unit cell problem with an arbitrarily shaped void at the center ($\Omega = [0,1] \times [0,1]$). The prior distribution of the void radius is given by: \(
r = 0.2 + \frac{0.2}{1+\exp(\bar{r})}\), with \(\quad \bar{r} \sim \mathcal{N}(0,4^2(-\nabla + 3^2)^{-1})\).
This distribution ensures $0.2 \leq r \leq 0.4$. The unit cell is clamped at the bottom edge, with a tensile traction $t = [0,100]$ applied at the top edge. The material is an incompressible Rivlin-Saunders material (Pascon, 2019) with the constitutive model 

\begin{equation}
\sigma = \frac{\partial w(\epsilon)}{\partial \epsilon}, \quad w(\epsilon) = C_1(I_1 - 3) + C_2(I_2 - 3)
\end{equation}
where $I_1 = \text{tr}(C)$ and $I_2 = \frac{1}{2}[(\text{tr}(C))^2 - \text{tr}(C^2)]$ are scalar invariants of the right Cauchy-Green stretch tensor $C = 2\epsilon + I$. The energy density function parameters are $C_1 = 1.863 \times 10^5$ and $C_2 = 9.79 \times 10^3$.
This benchmark is to predict the inner stress of the elasticity material based on the material structure, which is discretized into 972 points. The training set contains 1,000 samples, and the test set contains 200 samples.

\subsection{Plasticity}
In this case, we investigate a plastic forging problem where a material block $\Omega = [0, L] \times [0, H]$ is impacted at time $t = 0$ by a frictionless rigid die parameterized by a function $S_d \in H^1([0, L]; R)$, moving at constant velocity $v$. The block is clamped at the bottom edge while displacement boundary conditions are imposed on the top edge. The governing equations are identical to the Elasticity benchmark but with an elasto-plastic constitutive model

\begin{equation}
\begin{aligned}
\sigma &= \mathbf{C} : (\epsilon - \epsilon_p), 
\dot{\epsilon}_p &= \lambda \nabla_\sigma f(\sigma), 
f(\sigma) &= \sqrt{\frac{3}{2}} \left\| \sigma - \frac{1}{3}\text{tr}(\sigma) \cdot I \right\|_F - \sigma_Y
\end{aligned}
\end{equation}
where $\lambda \geq 0$ is the plastic multiplier satisfying $f(\sigma) \leq 0$ and $\lambda \cdot f(\sigma) = 0$. The stiffness tensor $\mathbf{C}$ has parameters: Young's modulus $E = \SI{200}{\giga\pascal}$, Poisson's ratio 0.3, yield strength $\sigma_Y = \SI{70}{\mega\pascal}$, and mass density $\rho^s = \SI{7850}{\kilo\gram\per\cubic\meter}$. The target solution operator maps from the shape of the die to the time-dependent mesh grid and deformation. The data is given on a 101 × 31 structured mesh with 20 steps time marching. 900 samples with different die shapes are used for model training and 80 new samples are for testing.
\subsection{Airfoil}
In this case, we consider the transonic flow over an airfoil, where the governing equation is the Euler equation
\begin{equation}
    \begin{aligned}
        \frac{\partial \rho}{\partial t} + \nabla \cdot (\rho \mathbf{v}) &= 0, \\
        \frac{\partial (\rho \mathbf{v})}{\partial t} + \nabla \cdot (\rho \mathbf{v} \otimes \mathbf{v}) + \nabla p &= \mathbf{0}, \\
        \frac{\partial E}{\partial t} + \nabla \cdot \left( (E + p) \mathbf{v} \right) &= 0
    \end{aligned}
\end{equation}
where \(\rho\) is the fluid density, \(\mathbf{v}\) is the velocity vector, \(p\) is the pressure, and \(E\) is the total energy per unit volume, with viscous effects neglected. This task is to estimate the Mach number based on the airfoil shape, where the input shape is discretized into structured mesh with shape 221 × 51 and the output is the Mach number for each mesh point. 1000 samples of different airfoil designs are used for training and another 200 samples for testing.
\subsection{Pipe}
In this case, we consider the incompressible flow in a pipe. The governing equation is the incompressible Navier-Stokes equation
\begin{equation}
\begin{aligned}
    \frac{\partial}{\partial t} (\mathbf{v}) + (\mathbf{v} \cdot \nabla)\mathbf{v} &= -\nabla p + \nu \nabla^2 \mathbf{v}, \\
    \nabla \cdot \mathbf{v} &= 0
\end{aligned}
\end{equation}
where $\mathbf{v}$ is the velocity vector, $p$ is the pressure, and $\nu = 0.005$ is the viscosity. The target is to estimate the horizontal fluid velocity given the pipe geometry. Each case discretizes the pipe into structured mesh with size 129 × 129. 1000 samples with different pipe shapes are used for model training and 200 new samples are for testing, which are generated by controlling the centerline of the pipe.
\subsection{Navier-Stokes}
We consider the two-dimensional Navier-Stokes equations in vorticity form for a viscous, incompressible fluid on the unit torus.
\begin{equation}
\begin{aligned}
\partial_{t}w(x,t) + u(x,t) \cdot \nabla w(x,t) &= \nu \Delta w(x,t) + f(x), && x \in (0,1)^{2}, \, t \in (0,T], \\
\nabla \cdot u(x,t) &= 0, && x \in (0,1)^{2}, \, t \in [0,T], \\
w(x,0) &= w_{0}(x), && x \in (0,1)^{2}
\end{aligned}
\end{equation}
where the velocity field $u \in C([0,T]; H_{\mathrm{per}}^{r}((0,1)^{2}; R^{2}))$ for any $r > 0$ is the velocity field, $w = \nabla \times u$ is the vorticity, $w_{0} \in L_{\mathrm{per}}^{2}((0,1)^{2};R)$ is the initial vorticity, $\nu \in R_{+}$ is the viscosity coefficient, and $f \in L_{\mathrm{per}}^{2}((0,1)^{2}; R)$ is the forcing function. We set viscosity as $10^{-5}$. The task is to predict the fluid in the next 10 steps based on the observations in the past 10 steps. 1000 samples with different initial conditions are generated for training, and 200 new samples are used for testing.
\subsection{Darcy}
We consider the steady-state of the 2-D Darcy flow equation in the unit box, which is a second-order, linear, elliptic PDE:
\begin{equation}
\begin{aligned}
-\nabla \cdot \big(a(x)\nabla u(x)\big) &= f(x), \quad && x \in (0,1)^2, \\
u(x) &= 0, \qquad && x \in \partial(0,1)^2
\end{aligned}
\end{equation}
with a Dirichlet boundary condition, where $a \in L^\infty((0,1)^2;{R}_+)$ is the diffusion coefficient and $f \in L^2((0,1)^2; R)$ is the forcing function.
The solution domain is discretized into a 421 × 421 regular grid. The simulation results are then downsampled into 85 × 85 resolution. 1000 samples are used for training and 200 samples are generated for testing, where different cases contain different medium structures.
\subsection{AirfRANS}
This dataset is mainly motivated by a realistic shape optimization problem. Each case is discretized into 32,000 mesh points. By changing the airfoil shape, Reynolds number, and angle of attack, AirfRANS provides 1000 samples, where 800 samples are used for training and 200 for testing. We need to estimate air velocity, pressure and viscosity for surrounding space and pressure is recorded for the surface. We focus on the lift coefficient estimation and the pressure quantity on the volume and surface, which is essential to the take-off and landing stages of airplanes.
\begin{figure*}[h]
\centering
\includegraphics[width=0.7\columnwidth]{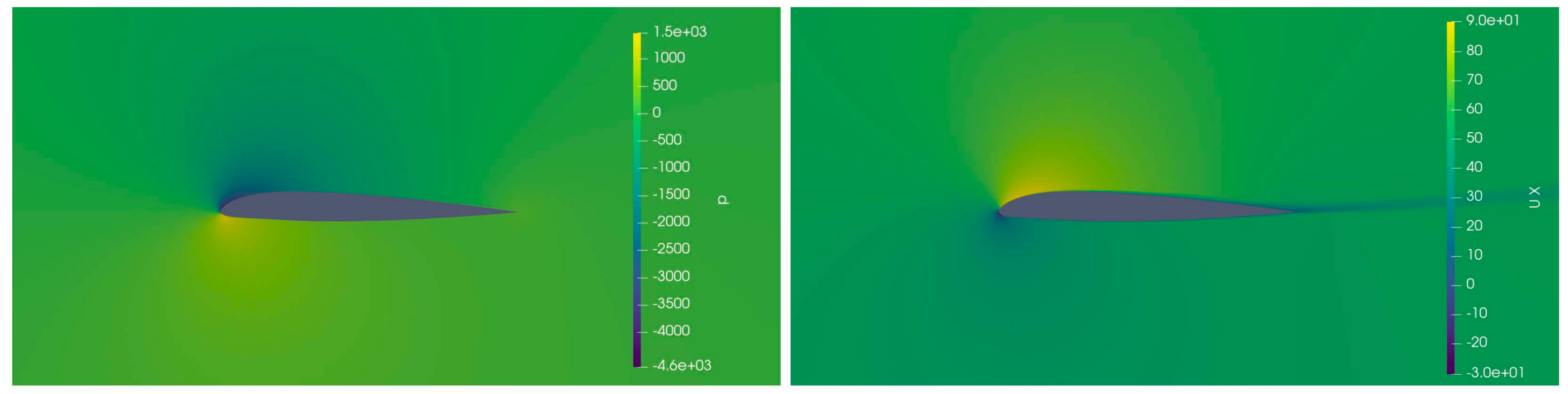} 
\caption{Visualization of a ground truth in AirfRANS. The left subfigure visualizes pressure field and the right visualizes $x$-component of the velocity field.}
\label{figAir}
\end{figure*}
\subsection{Shape-Net Car}
The Shape-Net Car dataset contains 889 samples, each representing a finite element solver simulation result. The simulations computed both the time-averaged fluid pressure on the surface and the velocity field around the car by solving the large-scale Reynolds-averaged Navier-Stokes (RANS) equations using the $k$-$\varepsilon$ turbulence model with SUPG (Streamline-Upwind/Petrov-Galerkin) stabilization. All simulations were conducted with a fixed inlet velocity of \SI{72}{\kilo\meter\per\hour} and a Reynolds number of $5 \times 10^6$. Afterward, we can calculate the drag coefficient based on these estimated physical fields.

\begin{figure*}[h]
\centering
\includegraphics[width=0.45\columnwidth]{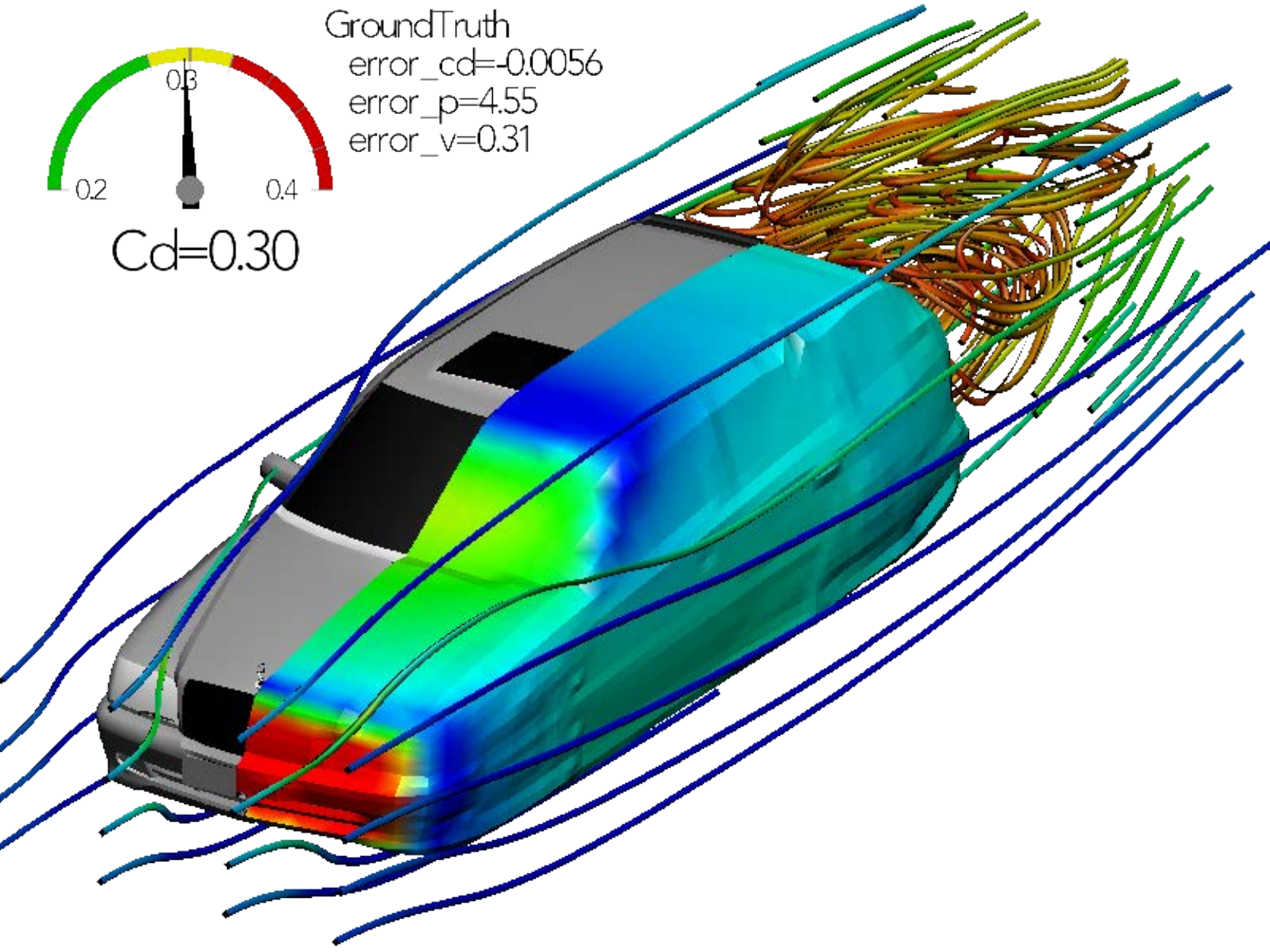} 
\caption{Visualization of a ground truth in the ShapeNet Car dataset.}
\label{figCar}
\end{figure*}

\section{Metrics}
\subsection{Relative L2}
The relative L2 error between the ground-truth physics field $u$ and the predicted field $\hat{u}$ is defined as:

\begin{equation}
\text{Relative L2 Error} = \frac{\| u - \hat{u} \|_{2}}{\| u \|_{2}} 
    = \frac{
        \sqrt{\int_\Omega |u(x) - \hat{u}(x)|^2 \, \mathrm{d}x}
    }{
        \sqrt{\int_\Omega |u(x)|^2 \, \mathrm{d}x}
    }
\end{equation}
\subsection{Relative L2 for drag and lift coefficients}
The relative L2 loss is defined between the ground truth coefficients and the ones calculated from the predicted velocity and pressure fields. For unit density fluid, the coefficient (drag or lift) is defined as
\begin{equation}
	\begin{split}
C=\frac{2}{v^2A}\left(\int_{\partial\Omega}p(\boldsymbol{\xi})\left(\widehat{n}(\boldsymbol{\xi})\cdot\widehat{i}(\boldsymbol{\xi})\right)\mathrm{d}\boldsymbol{\xi} +\int_{\partial\Omega}\tau(\boldsymbol{\xi})\cdot\widehat{i}(\boldsymbol{\xi})\mathrm{d}\boldsymbol{\xi}\right)
    \end{split}
\end{equation}
where $v$ is the speed of the inlet flow, $A$ is the reference area, $\partial\Omega$ is the object surface, $p$ denotes the pressure function, $\widehat{n}$ means the outward unit normal vector of the surface, $\widehat{i}$ is the direction of the inlet flow and $\tau$ denotes wall shear stress on the surface. As for the lift coefficient of AirfRANS, $\widehat{i}$ is set as $(0,0,-1)$ and for the drag coefficient of Shape-Net Car, $\widehat{i}$ is set as $(-1,0,0)$ and $A$ is the area of the smallest rectangle enclosing the car fronts.
\subsection{Spearman’s rank correlations}
Spearman’s rank correlation quantifies the capability of a model to rank designs effectively, enabling efficient identification of the optimal design based on predicted rankings. Given the ground truth coefficients $C=\{C^{1},\cdots, C^{K}\}$ and the model predicted coefficients $\widehat{C}=\{\widehat{C}^{1},\cdots,\widehat{C}^{K}\}$, Spearman’s correlation coefficient is defined as
\begin{equation}
	\begin{split}
\rho=\frac{\operatorname{cov}\left(R(C)R(\widehat{C})\right)}{\sigma_{R(C)}\sigma_{R(\widehat{C})}}
    \end{split}
\end{equation}
where $R$ is the ranking function that transforms raw data values into their corresponding ranks, $\operatorname{cov}$ denotes the covariance, and $\sigma$ represents the standard deviation of the rank variables.
\section{Extended Model Analysis}
\subsection{Full Results on Scaling Experiments} 
Figure~\ref{figwidth} shows relative L2 test error v.s. the width of GFocal. Evidently, when the model width $\leq$ 128, the reduction in error is positively correlated with the width. The observation that the model with a width of 256 exhibits higher relative L2 test error than a width of 128 is possibly attributed to excessive parameters in neural operators leading to overfitting. Considering the favorable balance between performance and computational cost, we adopt the model configuration with a width of 128. Combined with the two sets of experimental results on Elasticity, this demonstrates the potential of \textbf{GFocal} as a large-scale pre-trained neural operator model. 
\begin{figure*}[h]
\centering
\includegraphics[width=0.5\textwidth,      % 宽度缩放到文本宽度的 80%
  height=0.5\textheight,    % 高度同步缩放
  keepaspectratio,          % 保持原始宽高比
  draft=false,     ]{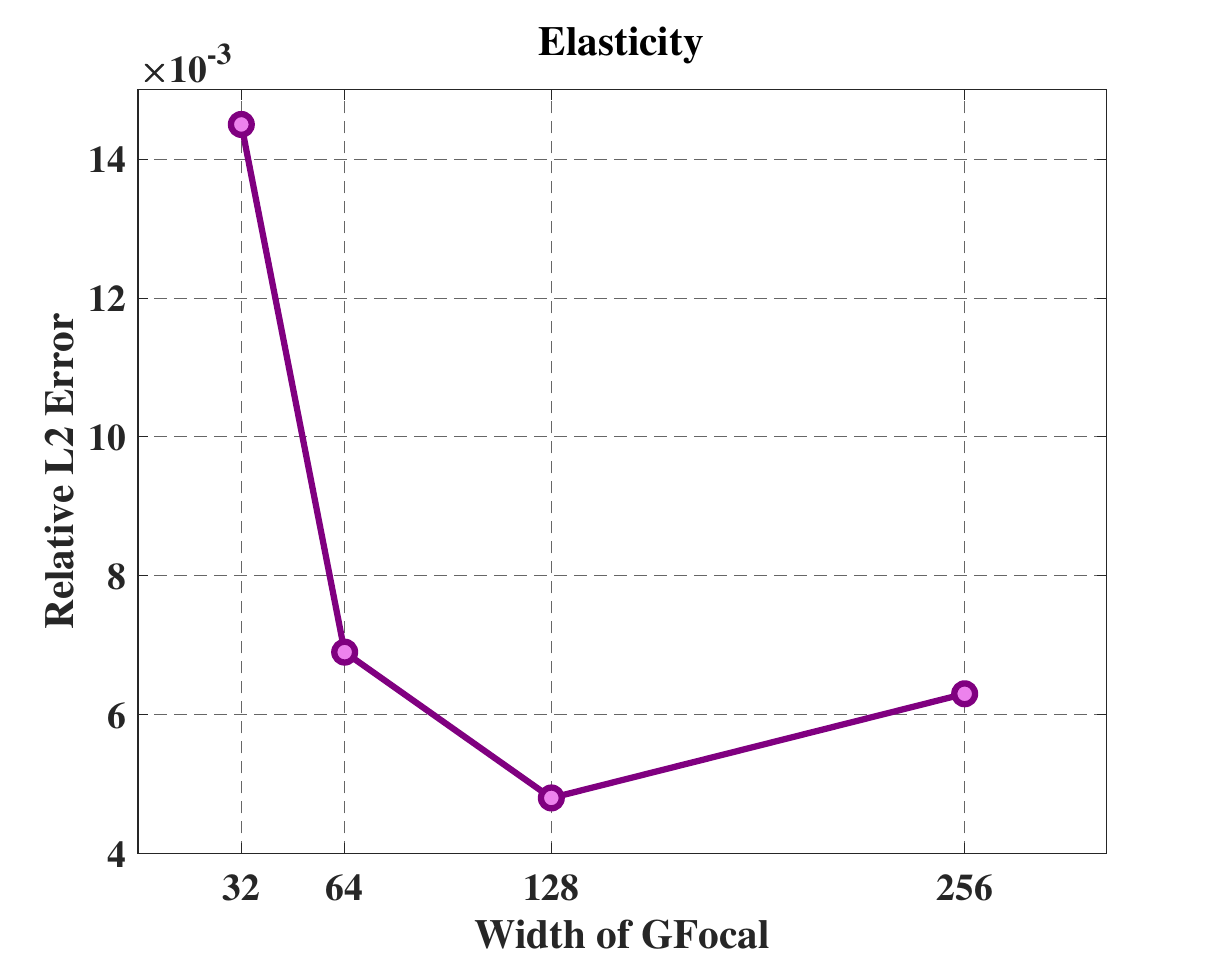}       % 确保实际图像显示]{width.png} 
\caption{Results of scaling experiments of GFocal on the Elasticity benchmarks.}
\label{figwidth}
\end{figure*}
\subsection{Ablation Study Visualization}
Figure~\ref{figablation} presents visualization to the ablation study. These visualizations provide valuable insights into the capability of global blocks to model global correlation of physical fields and the capability of focal blocks  to model complex physical details. Overall, the model demonstrates strong performance in modeling interactions of complex physical states. %Overall, our model demonstrates robust ability to learning physical details and interactions on complex geometries.

\begin{figure*}[h]
\centering
\includegraphics[width=1\columnwidth]{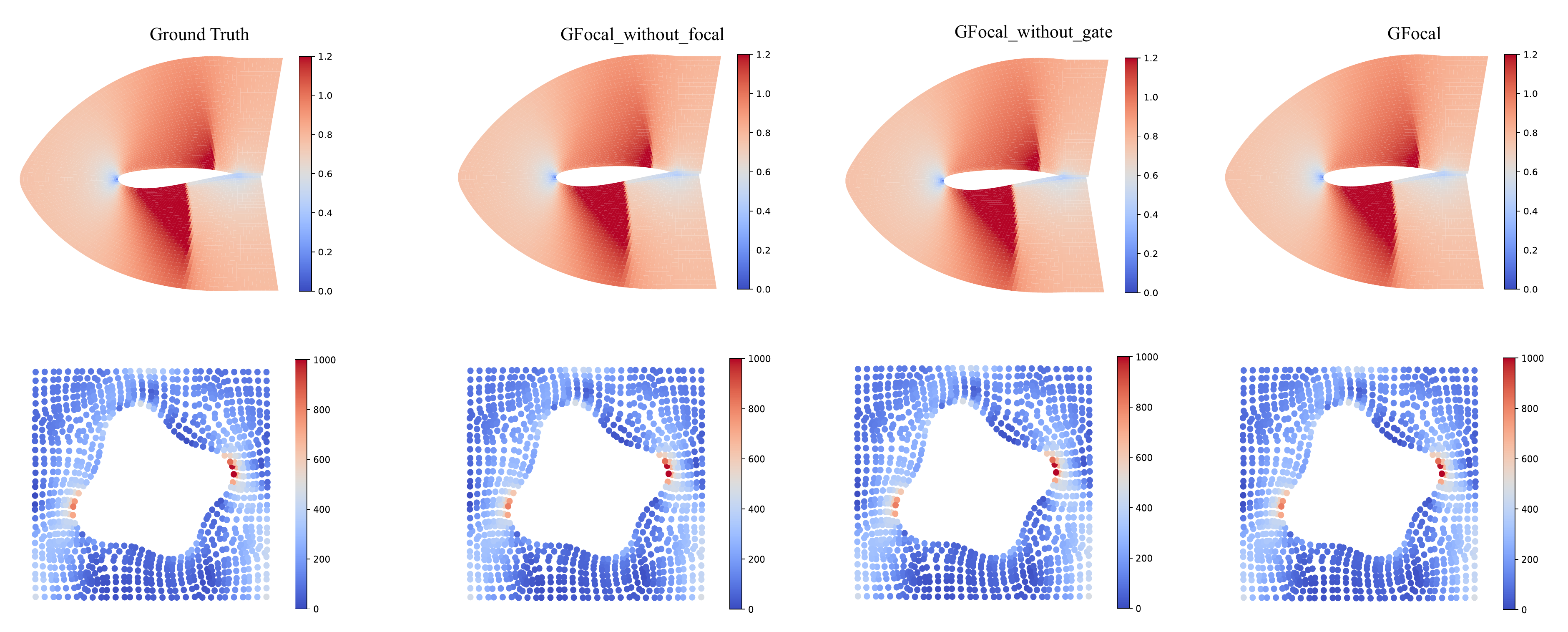} 
\caption{Ablation study visualization on the Airfoil and Elasticity benchmarks.}
\label{figablation}
\end{figure*}
\subsection{Implementations}
The implementations of GFocal for different benchmarks are summarized in Table~\ref{tab:training_model_detail}. $\mathcal{L}_{\mathrm{v}}$ and $\mathcal{L}_{\mathrm{s}}$ represent the loss on volume and surface fields respectively. As for Darcy, we adopt an additional spatial gradient regularization term $\mathcal{L}_{\mathrm{g}}$ following Transolver.

\begin{table}[h]
	\caption{Training configurations of  GFocal.}
	\label{tab:training_model_detail}
	\centering
		\begin{sc}
			\renewcommand{\multirowsetup}{\centering}
			\setlength{\tabcolsep}{2.2pt}
			\begin{tabular}{l|ccccc}
				\toprule
                \multirow{3}{*}{Benchmarks} & \multicolumn{5}{c}{Training Configuration (Shared in all baselines)} \\
                    \cmidrule(lr){2-6}
			  & Loss & Epochs & Initial LR & Optimizer & Batch Size \\
			    \midrule
                 Elasticity & & \multirow{5}{*}{500} & \multirow{5}{*}{$10^{-3}$} & & 2  \\
			Plasticity & & & &  & 8  \\
        	Airfoil & Relative & & & AdamW & 4 \\
                Pipe & L2 & & & & 4  \\
                Navier–Stokes & & & & & 2  \\
                 \midrule
                Darcy & $\mathcal{L}_{\mathrm{rL2}}+0.1\mathcal{L}_{\mathrm{g}}$ & 500& \multirow{3}{*}{$10^{-3}$}&AdamW & 2\\
               
                Shape-Net Car & $\mathcal{L}_{\mathrm{v}}+0.5\mathcal{L}_{\mathrm{s}}$ & 200 &  & \multirow{2}{*}{Adam} & 1\\
                AirfRANS & $\mathcal{L}_{\mathrm{v}}+\mathcal{L}_{\mathrm{s}}$  & 400 & &  & 1 \\
				\bottomrule
			\end{tabular}
		\end{sc}
\end{table}

\subsection{Hyperparameter Setting}
We conducted an extensive hyperparameter search. The final hyperparameter settings for the different benchmarks are summarized in Table~\ref{tab:hyperparameters}.
\begin{table}[h]
\centering
\caption{Hyperparameter settings of GFocal for different benchmarks.}
\label{tab:hyperparameters}
\begin{tabular}{l | cccc}
\toprule
\textbf{Benchmark} & \textbf{Global Depth (M)} & \textbf{Focal Depth (K)} & \textbf{Channels (C)} & \textbf{Slice Num (L)}\\
\midrule
Elasticity & 5 & 5 & 128 & 32\\
Plasticity  & 5 & 5 & 128&  32\\
Airfoil & 4 & 4 & 128  & 32\\
\midrule
Pipe & 4 & 4 & 128  & 32\\
Navier–Stokes & 4 & 4 & 256& 32\\
Darcy & 6 & 3 & 128 & 64\\
\midrule
Shape-Net Car & 5 & 5 & 256  &32\\
AirfRANS & 5 & 5 & 256 &32\\
\bottomrule
\end{tabular}
\end{table}

\subsection{Efficiency}
To further demonstrate the model efficiency, we compare it with ONO and Transolver. We record the GPU memory (GB) and the cost of training time per epoch in the following table in Elasticity, Airfoil and Navier-Stokes. As shown in the Table~\ref{tab:efficency} and Fig.~\ref{figeff}, GFocal surpasses ONO, but still requires higher memory demands and longer training time compared with Transolver. To enhance the model's practicality, we will further streamline our model in future work.
\begin{table}[h]
\centering
\caption{Efficiency comparison between different models. We set the hyperparameters of Transolver and ONO according to their official configurations. However, in the table, the batch size during testing are set the same for all three models, which are {2, 4, 2} for the three benchmarks respectively. Two metrics are measured: the cost of GPU memory and the cost of training time per epoch.}
\label{tab:efficency}
\begin{tabular}{ccccc}
\toprule
\textbf{Metrics}& \textbf{Model} & \textbf{Elasticity} & \textbf{Airfoil} & \textbf{Navier-Stokes} \\
\midrule
\multirow{3}{*}{Memory (GB)}&GFocal& 1.81 & 9.26 & 22.53 \\
 &Transolver  & 0.68 & 4.21 & 9.71\\
 &ONO & 2.32 & 12.91 & 34.79  \\
\midrule
\multirow{3}{*}{Time (s/epoch)}&GFocal&13.55&37.80&248.90\\
 &Transolver  & 7.07 & 20.97 & 127.18\\
  &ONO & 18.32 & 45.90 & 358.71  \\
\bottomrule
\end{tabular}
\end{table}

\begin{figure*}[h]
\centering
\includegraphics[width=0.5\textwidth,      % 宽度缩放到文本宽度的 80%
  height=0.5\textheight,    % 高度同步缩放
  keepaspectratio,          % 保持原始宽高比
  draft=false, ]{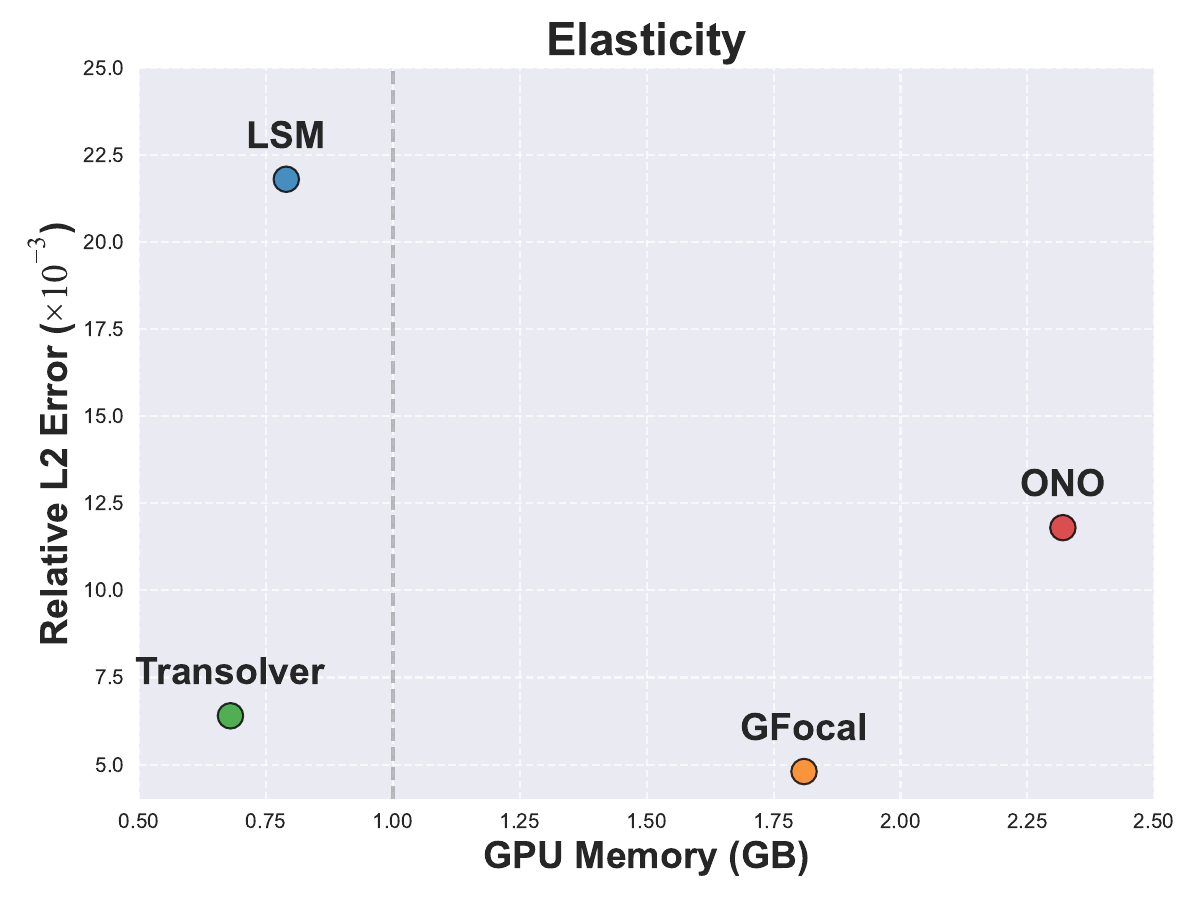} 
\caption{Efficiency comparison on Elasticity (972 mesh points). Metrics are measured with the batch size of 2.}
\label{figeff}
\end{figure*}

\end{document}